\documentclass[conference]{IEEEtran}
\IEEEoverridecommandlockouts
\usepackage{cite}
\usepackage{amsmath,amssymb,amsfonts}
\usepackage{algorithmic}
\usepackage{graphicx}
\usepackage{textcomp}
\usepackage[linesnumbered,ruled]{algorithm2e}
\usepackage{array}

\newlength\mylen
\newcommand\myinput[1]{%
  \settowidth\mylen{\KwIn{}}%
  \setlength\hangindent{\mylen}%
  \hspace*{\mylen}#1\\}
  
\newcolumntype{M}[1]{>{\centering\arraybackslash}m{#1}}
  
\def\BibTeX{{\rm B\kern-.05em{\sc i\kern-.025em b}\kern-.08em
    T\kern-.1667em\lower.7ex\hbox{E}\kern-.125emX}}
\begin{document}

\title{Unsupervised Learning using Pretrained CNN and Associative Memory Bank
{\footnotesize \textsuperscript{}}
\thanks{Partially supported by a TIRE grant from Louisiana Transportation Research Center.}
}

\author{\IEEEauthorblockN{Qun Liu}
\IEEEauthorblockA{\textit{Department of Computer Science} \\
\textit{Louisiana State University}\\
Baton Rouge, Louisiana \\
qliu14@lsu.edu}
\and
\IEEEauthorblockN{Supratik Mukhopadhyay}
\IEEEauthorblockA{\textit{Department of Computer Science} \\
\textit{Louisiana State University}\\
Baton Rouge, Louisiana \\
supratik@csc.lsu.edu}
}

\maketitle

\begin{abstract}
Deep Convolutional features extracted from a comprehensive labeled dataset, contain substantial representations which could be effectively used in a new domain. Despite the fact that generic features achieved good results in many visual tasks, fine-tuning is required for pretrained deep CNN models to be more effective and provide state-of-the-art performance. Fine tuning using the backpropagation algorithm in a supervised setting, is a time and resource consuming process. In this paper, we present a new architecture and an approach for unsupervised object recognition that addresses the above mentioned problem with fine tuning associated with pretrained CNN-based supervised deep learning approaches while allowing automated feature extraction. Unlike existing works, our approach is applicable to general object recognition tasks. It uses a pretrained (on a related domain) CNN model for automated feature extraction pipelined with a Hopfield network based associative memory bank for storing patterns for classification purposes. The use of associative memory bank in our framework allows eliminating backpropagation while providing competitive performance on an unseen dataset.

\end{abstract}


\section{Introduction}
 In the last few years, advances in  supervised \emph{Deep Learning} \cite{lecundeep} enabled by   Convolutional Neural Networks (CNN) \cite{krizhevsky2012imagenet} have given rise to  powerful  techniques for  solving a variety of problems in  Computer Vision \cite{nam2016learning,ciregan2012multi,gidaris2015object}, especially those involving  image classification and segmentation \cite{hariharan,hariharan1, deepsat, ieeetran,urtasun1,texture, character, texture1, ciregan2012multi,he2016deep,girshick2014rich,krizhevsky2012imagenet,chen2018deeplab}, visual tracking \cite{tracking}, etc. However, one of the bottlenecks faced by deep learning approaches based on CNN models trained using the backpropagation algorithm is the requirement of large amounts of labeled training data. Given that these models sometimes have billions of parameters, lack of  training data can result in overfitting  to the training dataset. While sophisticated regularization techniques \cite{Hastie} are today used to prevent overfitting, they cannot alleviate the need for availability of large amounts of labeled  training data. In many domains, acquiring large amounts of labeled training data can be prohibitively expensive or infeasible. To address the lack of large volumes of labeled training data, researchers have proposed zero-shot or one-shot approaches \cite{zeroshotRomero,ZeroshotNg,Oneshot}. In the one-shot approach \cite{Oneshot}, the authors use a Bayesian paradigm wherein  one  uses a prior probability distribution to represent  knowledge about categories of objects acquired apriori and then uses belief update to obtain a posterior distribution. Using this approach the authors have been able to recognize categories of objects based on five or fewer training examples.  However, unlike existing deep learning approaches, feature extraction in \cite{Oneshot} is not automated and depends identifying ``interesting regions'' in the image. In zero-shot learning \cite{zeroshotRomero,ZeroshotNg}, one needs a description of attributes characterizing the classes previously learned as well as information that relates them to unseen ones; given these inputs, zero-shot learning approaches can recognize unseen classes even without any training example.  In \cite{Vicarious}, using handcrafted features, the authors have been able to create generative models for character recognition  with little training data. However, the approach of \cite{Vicarious} can  not  be easily extended to general object recognition tasks.  
 
 There are two main reasons  why supervised deep learning approaches based on CNN models are hungry for labeled data. First, CNN-based object recognition approaches usually start from ``scratch'' without any prior knowledge about the object classes they are meant to recognize unlike zero-shot or one-shot approaches. In other words, no prior knowledge that was acquired while performing previous recognition tasks gets transfered to the new domain. On the other hand, CNN-based approaches  have the advantage of allowing  completely automated representation learning  as opposed to zero-shot or one-shot approaches.  The second reason for being data hungry is that deep learning approaches based on CNN models, during training,  use supervised learning based on the  backpropagation algorithm to estimate a large number of parameters (weights) based on training data. Such an algorithm demands a large number of  labeled training examples.  Unsupervised learning methods \cite{sabour2017dynamic,Unsupervised} have recently gained attention as a way of addressing the  labeled data-hungry nature of supervised deep learning approaches. However, existing unsupervised learning approaches have not been able to compete with supervised ones in terms of performance.  
 
\begin{figure*}[t!]
\centering
\includegraphics[width=18cm, height=10cm]{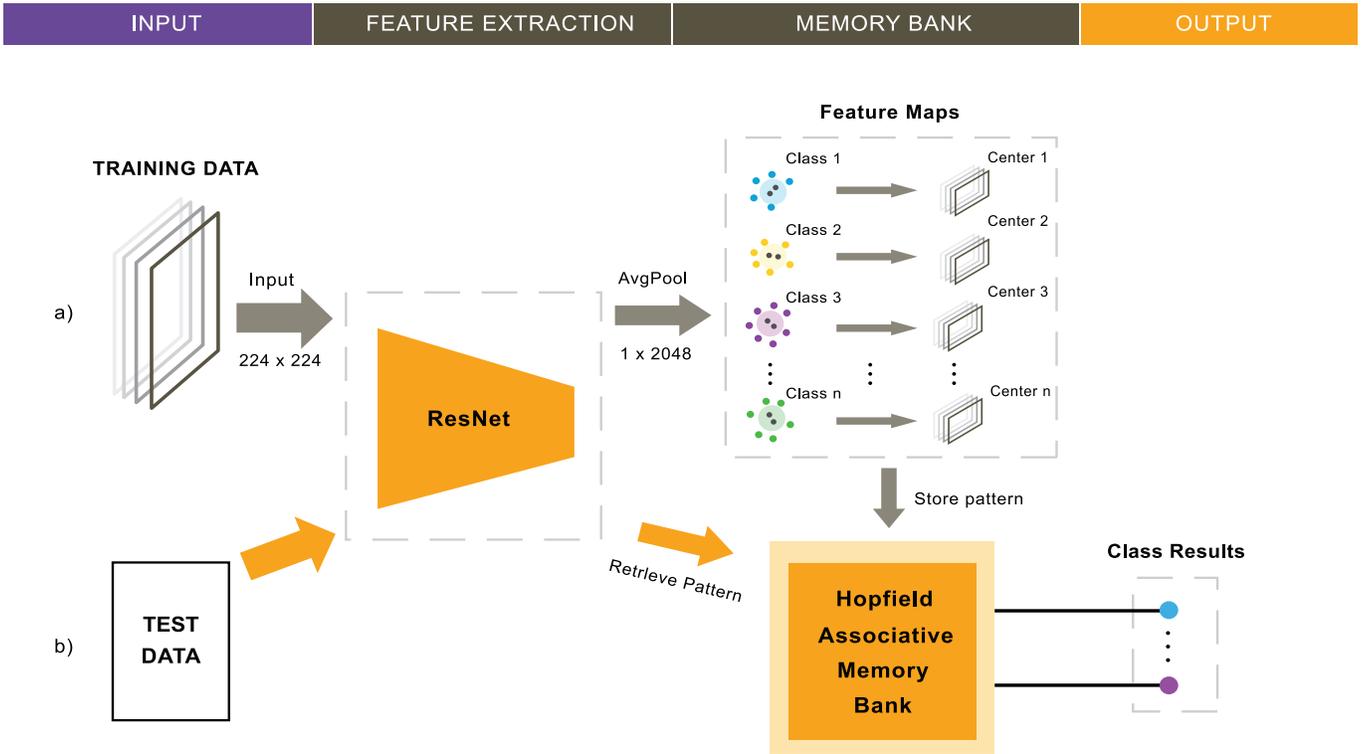}
\caption{Overview of Pipeline Framework Architecture.}
\label{fig:bf}
\end{figure*} 
 
 In this paper, we present a new architecture and an approach for  unsupervised object recognition that addresses the above-mentioned problems associated with CNN-based supervised deep learning approaches while allowing automated feature extraction unlike zero-shot and one-shot approaches. Unlike \cite{Vicarious}, our approach is applicable to general object recognition tasks. It  uses a pretrained (on a related domain)  CNN model for automated feature extraction while it pipelines a Hopfield network \cite{sabahi2017hopfield} based memory bank for storing patterns for classification purposes. Figure \ref{fig:bf} shows the architecture of our approach.

It is well known  that the map responses  pooled from the  different layers  of a CNN model yield  more  advanced and complex descriptors  as compared to  handcrafted ones \cite{ren2015faster}. Together with the rise of deep learning, transfer learning \cite{pan2010} has enjoyed  great success and  has played a vital role in obtaining  good feature representations from a pretrained CNN model \cite{yosinski2014transferable}. In domains  where acquiring a  large-scale  labeled training dataset is very hard and almost infeasible,  CNN models, pretrained on another large-scale dataset (e.g., ImageNet\cite{deng2009imagenet}) from a  related domain that does not contextually differ too much,  have achieved competitive effectiveness  after fine tuning. 

During inference, our  framework uses  a pretrained CNN classification model  (on ImageNet) to extract feature maps from the input images, then computes the centers of these maps as  patterns fed to  the pipelined memory bank (Hopfield Network) to infer the test image class.


Compared with previous work  \cite{karpathy2014large,tzeng2015simultaneous} demonstrating the effectiveness of transfer learning with  CNNs  for object classification tasks, our work distinguishes itself  in two ways. First, in existing transfer learning frameworks that use a pretrained CNN model,  one needs to fine tune the pretrained CNN  parameters on the new dataset to obtain acceptable recognition  accuracy. But fine tuning a network having  millions of parameters trained on large-scale dataset through back propagation is  time and resource consuming. The use of associative memory bank in our framework allows  eliminating  backpropagation while providing  good performance on an unseen dataset. On the Caltech101 dataset, our framework achieved an accuracy of 91.0\%,  on the Caltech256 dataset it obtained  an accuracy of 77.4\%,   while on the CIFAR-10 dataset it provided an accuracy of 83.1\%; in all three cases, the performance  surpassed or achieved that obtained using existing state-of-the-art (Section \ref{exp}). 


Recently some studies \cite{krotov2016dense,krotov2017dense} demonstrate the potential of neural associative memory in pattern recognition and  its robustness to adversarial inputs. However, no previous work has tried to combine pretrained CNN models with an associative memory bank in an unsupervised setting. 


To the best of our knowledge, this is the first work that  provides an unsupervised  framework  that combines transfer learning  together with an associative memory bank,  uses a pretrained CNN model for automatically extracting features and is able to provide good performance in an unseen domain without fine tuning using the backpropagation algorithm. This paper makes the following contributions.  

\begin{itemize}
\item It addresses the two main reasons for the data-hungry nature of supervised deep learning approaches based on the CNN model:  (a) lack of transfer of prior knowledge to a new domain (b) the use of supervised learning using the backpropagation algorithm to estimate a large number of parameters (weights) based on training data.  It provides a  pipelined  unsupervised learning framework  that combines transfer learning  together with an associative memory bank,  uses a pretrained CNN model (on a related domain) for automatically extracting features and is able to provide good performance in an unseen domain without fine tuning using the backpropagation algorithm.
\item It experimentally demonstrates the effectiveness of the framework on the Caltech101, Caltech256, and CIFAR-10 benchmark datasets.
\end{itemize}\par

The paper is organized as follows. In Section \ref{rw}, we will present the related work briefly. The pipeline framework architecture and Hopfield Network will be discussed in Section \ref{pm}, the classification algorithm is provided. We will demonstrate the experimental results in Section \ref{exp} and the conclusions presented in Section \ref{con}.  

\section{Related Work}\label{rw}

Recently, supervised deep learning have been successfully applied to computer vision problems. The main reason for the success of deep learning  is the availability of  large volumes of labeled training data as well as computing power \cite{krizhevsky2012imagenet}. Researchers have come up with a plethora of   deep learning techniques and the accuracy on object recognition tasks has continued to improve over time already surpassing human-level performance \cite{he2015delving,mnih2015human}. Deep features learned from  pretrained CNN models have shown competitive performances  in vision related tasks. There has been a significant amount of work in reusing  deep features for reducing data requirement  in many  domains \cite{girshick2015fast,girshick2014rich}  like image classification, segmentation, object recognition, etc.\par

  In the real world, acquiring large volumes of  labeled  training data is very expensive and impractical at some contexts. Unsupervised learning is getting more and more  attention  since it  allows learning from unlabeled data \cite{masci2011stacked}. With a small labeled dataset, one can combine  labeled and unlabeled data in a semi-supervised  setting \cite{kingma2014semi}.  K-means is a popular clustering approach in  the unsupervised learning and is used as a part of  several deep  learning approaches \cite{gong2014compressing,ke2016fast,wang2015semantic}.\par

  Weightless Neural Network (also known as n-tuple or RAM networks) such as WISARD neural network\cite{aleksander1984wisard}, mimic the synaptic  activity in the brain. They store and recognize patterns that arise within the neuron  and not from the weights  on the connections. 
  In our  framework, we used a Hopfield network \cite{hopfield1984neurons}, a  recurrent neural network, as an associate memory bank. However, other types of auto-associative memory could also be considered in our framework.\par

The CapsNet architecture has been proposed recently by Sabour et. al. \cite{sabour2017dynamic}. It based on the notion of capsules, collections of neurons that play the role of nonterminals in a parse tree. The base layers of CapsNet computes activities from image pixel intensities that are input to the primary capsules. The output of a capsule is input to another in a higher layer with the target determined dynamically. A capsule represents an object class. During inference, the size of the output generated by a capsule indicates whether an object of the class represented by the capsule is possibly present in the test image. In our pipeline framework, core patterns used instead to characterize object classes and are stored in the associative memory bank as vectors.\par
In \cite{george2017generative}, the authors proposed  a probabilistic generative model   and  introduced a hierarchical model named Recursive Cortical Networks (RCN) that  handles the recognition, segmentation, and reasoning in a unified way  models objects using a combination of contours and surfaces. But in our  framework we  compute representations  using the pretrained CNN model in the first stage.

\section{Proposed Method}\label{pm}
In this section, we present the overview of our proposed method, then next we demonstrate the Memory Bank used in our pipeline framework. The core patterns selection and the Hopfield network described in Section \ref{cps} and Section \ref{hn}.

\subsection{Overview}\label{ov}
Our framework is designated to integrate an associative memory  bank with  a pretrained CNN model in an unsupervised setting to eliminate fine tuning using labeled training data for unseen object classes using  the backpropagation algorithm, while retaining state-of-the-art performance. The overall architecture of our framework is provided in Fig.1. As shown in the figure, we use a pretrained CNN model for feature embeddings extraction. Here a ResNet-50 \cite{he2016deep},  pretrained on ImageNet\cite{deng2009imagenet}, provides the basis for our framework.  Our framework uses the features  extracted from pool5 (of ResNet-50), before the dense layers, as the representations of the input images. During the training phase, our  framework calculates the class-specific features set, which is a set of all image features.   \emph{Core patterns}  are subsequently computed from the class-specific features set. We discuss  the details of this procedure in Section \ref{cps}. The set of core patterns  are stored in  the associative memory bank provided by the  Hopfield Network. During inference, we extract features of each test image from pool5, and then  compute patterns from input features that are used to retrieve  associated core patterns in  the memory bank and return the label for the test image.  More details are provided  in  Section \ref{pf}.

\subsection{Memory Bank}\label{mb}
We propose the notion of  an Associative memory bank for storing the core patterns.  The associative memory based on a  Hopfield network  allows retrieving patterns during the inference  stage and storing patterns during training stage. Given training images having  $n$ classes with $m$ images for class $i$, we have $p_i=\{p_i^1, p_i^2,...,p_i^m\}$ \emph{memory patterns}  computed from the features $f$ pooled from pool5 in the pretrained CNN model. Core patterns  are then computed based on  sets of memory patterns and stored in the memory bank. Details of computation of core patterns  are presented in the next subsection.

\begin{figure}[t!]
    \centering
    \includegraphics[height=5.3cm]{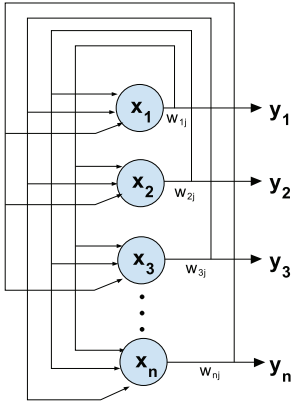}
    \caption{Hopfield network structure.}
    \label{fig:hp}
\end{figure}

\subsection{Core Pattern Computation and Storage}\label{cps}
From a set of  patterns $p_i=\{p_i^1, p_i^2,...,p_i^m\}$, that denotes the set of patterns for $i$-th class, extracted from pool5 in the pretrained CNN model, core patterns are computed.   The computation of the core patterns uses the  K-means algorithm, an unsupervised clustering approach, for calculating the cluster centers $c_i=$\{$c_i^1$,$c_i^2$,...,$c_i^k$ $\mid$ $k$$\in$1,2,...,m\}, as core patterns for each set $p_i$, using  Euclidean distance metric.  One or more core pattern can be created for each class.  The set of core patterns $c = \cup_{i=1}^n c_i$ is stored in the associative memory bank during the training stage. These core patterns in the memory bank serve as the candidates for   retrieval during inference. When a test image feed to the network, its corresponding core pattern  is retrieved from the  memory bank and its associated class  is  subsequently determined.  

\subsection{Hopfield Network}\label{hn}
The Hopfield associative memory is a single layer fully connected recurrent neural network, shown in Fig.\ref{fig:hp}.  The neurons in  a Hopfield network can be updated either  asynchronously or synchronously. For the asynchronous case, a neuron gets updated in a random or fixed order once its weighted input sum  is calculated, whereas in the synchronous case, all neurons get updated at same time. Given a network with $N$ neurons, the weighted input sum of a neuron, known as local field, can be described by the following equation.
\begin{equation}
\xi_{i}= \sum_{j = 1}^{N}w_{ij}x_j \label{eq1}
\end{equation}
where $i,j\in 1,2,...,N$, the synaptic weight $w_{ij}$ is  the connection weight between the $i$-th output and $j$-th input, and $x_j$ is the state of $j$-th input. The state of the entire network can be represented by a vector $v=\left[x_1,x_2,...,x_{N}\right]$.  Each  $N$ dimensional input pattern $\varphi$  can be  represented by $N$ neurons in  a Hopfield network. 

A Hopfield network memorizes $z$ core patterns  denoted by $\varphi_{1},\varphi_{2},...,\varphi_{z}$ during training stage, and retrieve the stored patterns during inference. Hebbian learning  is used for memorizing the patterns by the  Hopfield network. The connection weights $w_{ij}$ are computed by the following equation.

\begin{equation}
 w_{ij} =\begin{cases} \frac{1}{N} \sum_{u=1}^{z}\varphi_{u,i}^{} \varphi_{u,j}^{}  & i \neq j\\0 & i = j\end{cases} \label{eq2}
\end{equation}
 where $\varphi_{u,i}$ and $\varphi_{u,j}$ are the responses (states) to the pattern $\varphi_u$ of the $i$-th and the $j$-th neurons. 
Note that the weight connection of $i$-th and $j$-th neuron is symmetric, i.e.,  $w_{ij}=w_{ji}$.\par

\begin{figure}[t!]
    \centering
    \includegraphics[width=8cm]{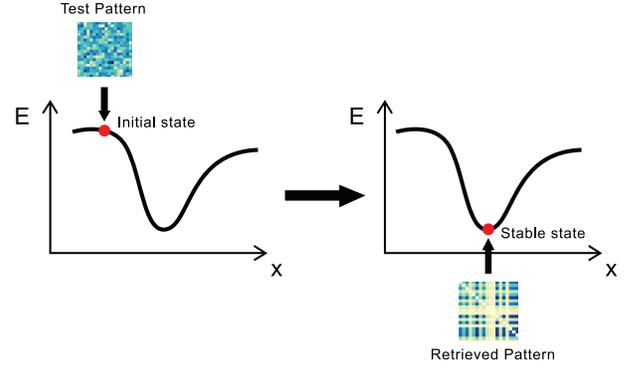}
    \caption{Pattern retrieve based on Hopfield Network.}
    \label{fig:state}
\end{figure}

For retrieving, as shown in Fig.\ref{fig:state}, suppose  that we have a test pattern $\varphi_{test}$,  
then $x_i$ is the state of the $i$-th element in test pattern, denotes as $\varphi_{test,i}$ and $i=1,2,...,N$.
Then all the neurons  update their state asynchronously as described by the following equation,
\begin{equation}
x_i(t+1)=\it sign(\sum_{j=1}^{N} w_{ij}x_j(t)) \label{eq3}
\end{equation}
where $x_i(t)$ is the state of the $i$-th neuron at time $t$. The update takes the network  to a ``lower energy'' state.   Once the energy of the network is minimized, it stabilizes. Following  equation \eqref{eq1} the energy $E_i$ for neuron $i$ can be described by the following equation,
\begin{equation}
E_i=-\frac{1}{2}\xi_{i}x_i \label{eq4}
\end{equation}

Then the energy for the entire network \cite{sabahi2017hopfield} can be calculated using  the following equation,
\begin{equation}
E\left (v\right )=\sum_{i=1}^{N}E_i=-\frac{1}{2}\sum_{i=1}^{N}\sum_{j=1}^{N}w_{ij}x_ix_j \label{eq5}
\end{equation}

Even  if the test pattern is incomplete or broken, a Hopfield network can still retrieve the corresponding stored  core pattern  from  memory due to its intrinsic of error-correcting and noise-resilience property. It can be shown that the network will converge to a stable state, once the energy of the network  minimizes and reaches the energy minimum or attractor basin (see Figure \ref{fig:state}). The associated final states in the network then represent the core patterns associated with  the test pattern. 

\subsection{Classification Algorithm}\label{pf}
We use a  Hopfield network as associative memory to store and retrieve core patterns. For a classification problem, after  retrieving  a pattern, the associated class is computed. The problem therefore transforms to a class matching one.  The classification algorithm is provided in Algorithm  \ref{alg1}. \par

The patterns memorized in the Hopfield network can be characterized by  the weight matrix \cite{ladwani2017hopfield} and can be retrieved using equation  \eqref{eq2}. In the following, for a pattern $T$, we denote its weight matrix by $W_T$. 


To obtain the retrieved pattern(s) by matching the test pattern with the stored patterns by  computing and comparing similarities among their weight matrix, we use the following  distance metric between the weight matrices as a similarity measure,
\begin{equation}
\mathit{Diff}(T,S)=\sqrt{\sum_{i=1}^{N}\sum_{j=1}^{N}(W_{T_{ij}}-W_{S_{ij}})^2} \label{eq7}
\end{equation}
where $T$ and $S$ are the test pattern and stored  core pattern respectively and $N$ is the dimension of the weight matrices. Then the stored  core pattern can be  retrieved  based on the following equation.
\begin{equation}
{\cal R} =argmin_{\varphi^i_k \mid i\in 1..z,k \in 1..n} \; \mathit{Diff}(T,\varphi^i_k) \label{eq8}
\end{equation}
where $\varphi^i_k$ is the $i$-th stored  core pattern for the $k$-th class.  Thus the stored core  pattern(s) that have the minimal distance from the test pattern is  retrieved.  \par

For each test pattern $t$ in  a test image, the algorithm computes the difference between $t$ and  each stored core pattern $\varphi^i_k$, $i \in \{1,\ldots, z\}$, $k \in\{1,\ldots, n\}$  in  class  $k$ using  the formula in equation \eqref{eq7}. 
Then  equation \eqref{eq8} is used to acquire   the set of patterns $R$  that has the minimal distance from $t$. If $R$ is a singleton, then the class label $l$ associated with its unique pattern is returned  as the result of  the classification. If equation \eqref{eq8} results in multiple core patterns, the label of one of them  is randomly chosen as the label of the test pattern. 

\begin{algorithm}
\SetKwInOut{Input}{Input}
\SetKwInOut{Output}{Output}
\caption{Classification Algorithm} \label{alg1}
 \Input{Test pattern $t$\;}
 \myinput{\ \ Stored patterns $\varphi^i_k$, $i \in 1..z$, $k \in 1..n$.}
 \Output{Class label $l$  of test pattern.}
 Initialize ${\cal R} \gets \emptyset$ \; \tcp{Initialize ${\cal R}$ to empty list}
${\cal R} =argmin_{\varphi^i_k \mid i\in 1..z,k \in 1..n} \; \mathit{Diff}(t,\varphi^i_k)$\;\tcp{${\cal R}$ is a list of patterns}
  \eIf{$len({\cal R}) > 1$}{
   $\varphi = random\_choice({\cal R})$\;
   \Return $l = label(\varphi)$\;
   }{
   \Return $l = label(\mathit{first}({\cal R}))$\;
  }
\end{algorithm}

\section{Experimental Evaluation}\label{exp}
To  illustrate the performance of our  framework with respect to  classification, we conducted  experiments presented in this section. The experiments   focus on three popular object classification datasets: Caltech101\cite{fei2007learning}, Caltech256\cite{griffin2007caltech}, and CIFAR-10\cite{krizhevsky2009learning}. We start with the  details of these datasets and then we contrast the results obtained using our framework with the state-of-the-art. We used ResNet-50 and VGG-16 as the pretrained CNN model in our  framework.

\subsection{Datasets}
\textbf{Caltech101} consists of 9144 images of 101 object categories and 1 background category. The variety of classes include faces, animals, camera, etc. The images in the dataset vary in the degree of shape and scale. For each category, it has about 40 to 800 images and most categories contain around 50 images.\par
\textbf{Caltech256} contains 30607 images for 256 object categories and 1 clutter category. It has  a minimum of 80 images for each category, compared to Caltech101, Caltech256 is more complex and challenging as it has more variety in the size, background, etc.\par
\textbf{CIFAR-10} contains ten classes with a total of  60000 RGB images with  image size  32x32 and each class has 6000 images.  1000 images from each class have  been randomly selected creating a  test dataset of 10000 images and the remaining were used for  training. The ten classes include airplane, automobile, bird, cat, etc.

\subsection{Implementation details}
 We used ResNet-50\cite{he2016deep} and VGG-16\cite{simonyan2014very} as pretrained CNN models (on ImageNet)  for feature extraction. The experiments mainly concentrate on the ResNet-50 model due to its state-of-the-art performance obtained for classification tasks and its  feature representations, but  results based on VGG-16 model  are also provided. Both models  have five convolutional blocks and the features are  pooled from the pool5, which has 1x1x2048 dimensions and 1x1x256 dimensions in ResNet-50 and VGG-16, respectively. The input image size for ResNet-50 and VGG-16 are same: 224x224. For unsupervised learning, to acquire the core patterns, K-means was used on the  memory patterns which are pooled features from the pretrained CNN model (see section \ref{cps}). 
 We followed the standard practice \cite{bo2013multipath,he2014spatial} for the experiments on Caltech101 and Caltech256 datasets:  for Caltech101,   training was performed on 30  randomly sampled images  and testing was done on  50 randomly selected images   per category  or on all images  for  those categories that had less  than 50 images; for Caltech256,  training was performed on  60  randomly sampled images  while the rest were reserved for testing. We did not divide the  CIFAR-10 dataset; we used its training data with no data augmentation in our  framework for training and its test data  was used for inference. 
 Recent efforts like \cite{snoek2015scalable}  used data augmentation  to achieve  very high accuracy in object classification.  In our experiments, we did not use data augmentation  approaches. We only resized the images for these datasets to the input size 224x224 for  the pretrained CNN model. The evaluation metric we adopted for the performance evaluation of our  framework is the average of the per-class accuracies obtained on the datasets.

\subsection{Evaluation}
\begin{figure}[htbp]
    \centering
    \includegraphics[width=8cm]{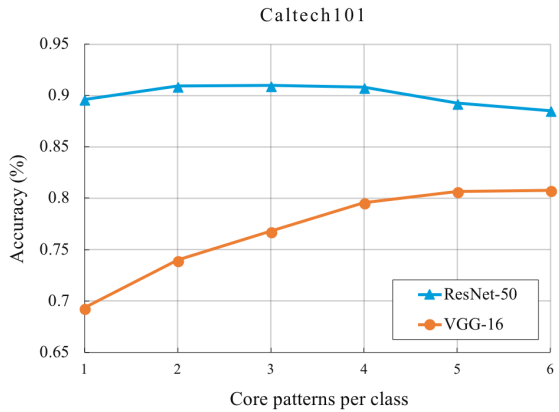}
    \caption{Impact of number of core patterns per class on Caltech101 dataset.}
    \label{fig:result2}
\end{figure}

\begin{figure}[htbp]
    \centering
    \includegraphics[width=8cm]{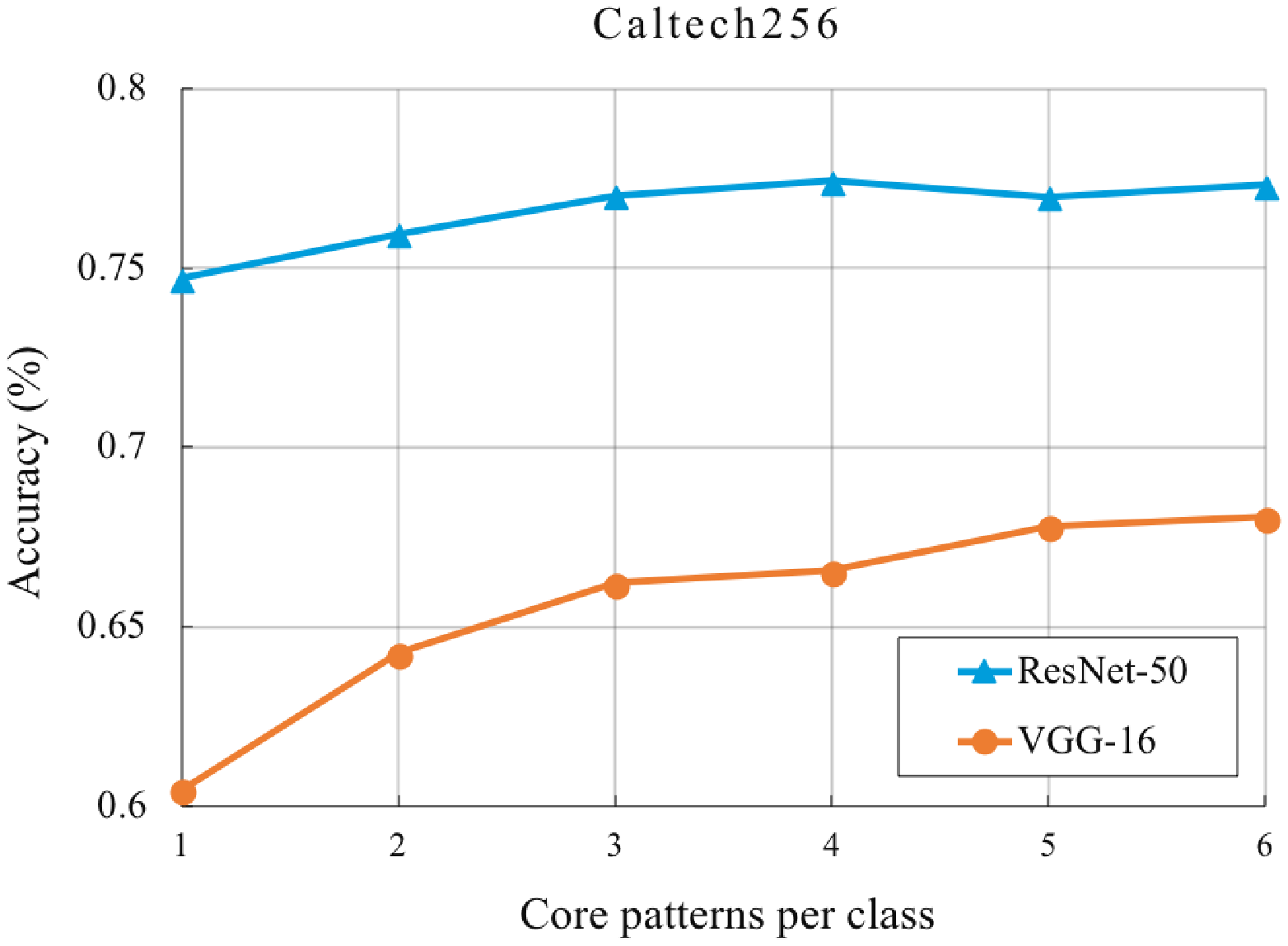}
    \caption{Impact of number of core patterns per class on Caltech256 dataset.}
    \label{fig:result3}
\end{figure}

\begin{figure}[t!]
    \centering
    \includegraphics[width=8cm]{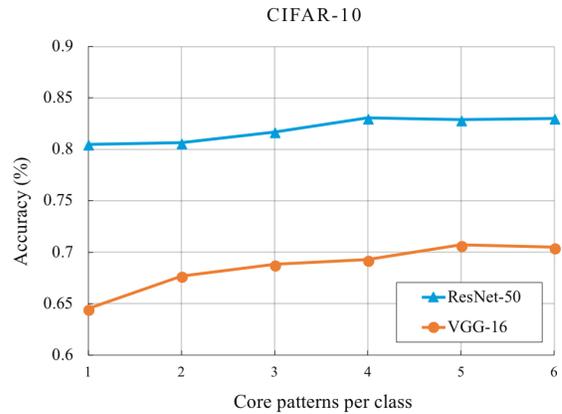}
    \caption{Impact of number of core patterns per class on CIFAR-10 dataset.}
    \label{fig:result4}
\end{figure}

\begin{figure}[htbp]
    \centering
    \includegraphics[width=8cm]{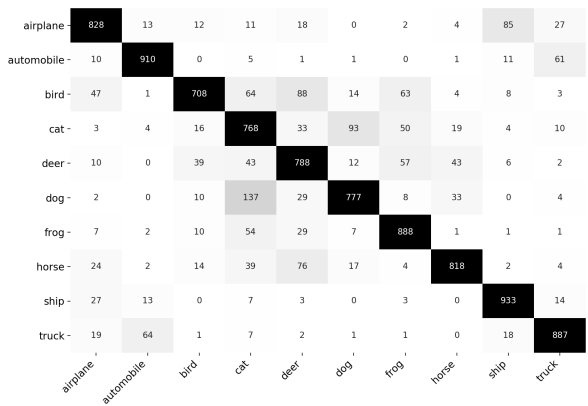}
    \caption{Confusion matrix for the CIFAR-10 dataset.}
    \label{fig:cifarcm}
\end{figure}

We  first evaluate the performance of our  framework with respect to  the number of core patterns. Recall that the core patterns are the centers computed by  the K-means algorithm. The results shown in Fig.\ref{fig:result2}, Fig.\ref{fig:result3}, and Fig.\ref{fig:result4}  are based on the Caltech101, Caltech256, and CIFAR-10 datasets from ResNet-50 and VGG-16 models. It is clear that from Fig.\ref{fig:result2} the  framework exhibited good performance and the classification accuracy improved with the increase in the number of core patterns for each class, but performance remained  relatively stable or  slight diminished after some point. It also showed similar behavior on Caltech256 and CIFAR-10 datasets but  relatively  more stable after increasing initially. This validates that multiple core patterns used in  framework can help in  performance improvement. 
The confusion matrix for CIFAR-10 dataset  is shown in Fig.\ref{fig:cifarcm}.  Most confusions arose from  inherently ambiguity  rather than the failure of the framework. We can see from the confusion matrix that the framework is confused in distinguishing between  dog and cat.  There is some confusion in distinguishing between truck and automobile.  There is very few or almost no confusion in distinguishing between  cat with automobile or truck. Since Caltech101 and Caltech256 datasets have a large number  of categories compared  to  the CIFAR-10 dataset, instead of computing the confusion matrix, we calculated the  number of false positives  for Caltech101 and Caltech256 datasets which are 236 and 1856, respectively.  

\subsection{Comparison with state-of-the-art methods}
In this section, we compare our classification results with the state-of-the-art methods based on  the Caltech101, Caltech256, and CIFAR-10 datasets. We show the comparisons in Table \ref{tab:cs1} and Table \ref{tab:cs2}. The classifiers for SVM and Softmax in the table are added on the top of the seven layers fixed model for which  pretraining was performed  on  the ImageNet dataset \cite{zeiler2014visualizing} and then retrained on the new corresponding training datasets.

\begin {table}[b!]
\caption {Comparison of classification accuracy (\%) of various methods on Caltech101 and Caltech256 datasets}
\begin{center}
\begin{tabular}{ |p{2.5cm}||M{2.0cm}|M{2.0cm}| }
 \hline
 Methods& Caltech101& Caltech256\\
 \hline
 Image\ Codes\cite{kuang2015hardware} & 71.4& 35.7 \\
 Shaban\cite{shaban2013local}   & 75.1& - \\
 SHDL\cite{singh2017scatternet} & 81.5 & - \\
 FL+EN\cite{zhu2014submodular}   & 83.2& - \\
 Softmax\cite{zeiler2014visualizing}   & 85.4& 72.6 \\
 SVM\cite{zeiler2014visualizing}   & 85.5& 71.7 \\
 Zeiler-Fergus\cite{zeiler2014visualizing} &86.5& 74.2 \\
 \hline
 Ours (VGG-16)&80.8& 68.1 \\
 Ours (ResNet-50)&\textbf{91.0}& \textbf{77.4} \\
 \hline
\end{tabular}
\label{tab:cs1}
\end{center}
\end{table}

The classification accuracies  achieved by our model  are \textbf{91.0\%} on Caltech101 and \textbf{77.4\%} on Caltech256. The classification accuracies of methods that we compared with  on CIFAR-10 are all reported with no data augmentation. Our model yielded an accuracy of \textbf{83.1\%} on CIFAR-10. While, \cite{springenberg2014striving,lee2016generalizing} provide better results than that obtained using our framework, they used supervised learning approaches in comparison to our unsupervised approach.  For CIFAR-10, our approach out performed  other unsupervised frameworks like DCGAN \cite{radford2015unsupervised}. While the unsupervised approach of \cite{exemplar1} out performed ours on CIFAR-10, unlike ours it used data augmentation. Though the approach of \cite{catg} outperforms ours on CIFAR-10, it is semi-supervised as opposed to ours being an unsupervised one. 


In our  framework, even if only one core pattern per class is used, the accuracy obtained is still  competitive: \textbf{89.6\%} on Caltech101, \textbf{74.7\%} on Caltech256, and \textbf{80.5\%} on CIFAR-10.

\begin {table}[t!]
\caption {Comparison of classification accuracy (\%) of our framework with existing techniques  on CIFAR-10 dataset with no data augmentation}
\begin{center}
\begin{tabular}{ |p{3.0cm}||M{2.0cm}| }
 \hline
 Methods& Accuracy\\
 \hline
 McDonnell et al.\cite{mcdonnell2015enhanced}   & 75.9 \\
 PCANet-2\cite{chan2015pcanet}   & 78.7 \\
 Cuda-convnet\cite{Krizhevsky:2014:Misc} & 82.0 \\
 CKN-CO\cite{mairal2014convolutional} & 82.2 \\
 DCGAN \cite{radford2015unsupervised} & 82.8 \\
 EX-CNN \cite{exemplar1} & 84.3 \\
 Conv-CatGAN \cite{catg} & 90.6 \\
 Springenberg et al.\cite{springenberg2014striving} & 90.9 \\
 Lee et al.\cite{lee2016generalizing} & \textbf{92.4} \\
 \hline
 Ours (VGG-16)&70.7 \\
 Ours (ResNet-50)&83.1 \\
 \hline
\end{tabular}
\label{tab:cs2}
\end{center}
\end{table}

\section{Conclusion}\label{con}
This paper proposed a framework that combines a  pretrained CNN model for feature extraction and  with  a Hopfield network as  an associative memory bank
to provide an unsupervised learning framework that provides competitive performance on benchmark datasets. 

\bibliographystyle{IEEEtran}
\nocite{Krizhevsky:2014:Misc}
\bibliography{main.bib}

\begin{thebibliography}{10}
\providecommand{\url}[1]{#1}
\csname url@samestyle\endcsname
\providecommand{\newblock}{\relax}
\providecommand{\bibinfo}[2]{#2}
\providecommand{\BIBentrySTDinterwordspacing}{\spaceskip=0pt\relax}
\providecommand{\BIBentryALTinterwordstretchfactor}{4}
\providecommand{\BIBentryALTinterwordspacing}{\spaceskip=\fontdimen2\font plus
\BIBentryALTinterwordstretchfactor\fontdimen3\font minus
  \fontdimen4\font\relax}
\providecommand{\BIBforeignlanguage}[2]{{%
\expandafter\ifx\csname l@#1\endcsname\relax
\typeout{** WARNING: IEEEtran.bst: No hyphenation pattern has been}%
\typeout{** loaded for the language `#1'. Using the pattern for}%
\typeout{** the default language instead.}%
\else
\language=\csname l@#1\endcsname
\fi
#2}}
\providecommand{\BIBdecl}{\relax}
\BIBdecl

\bibitem{lecundeep}
Y.~LeCun, Y.~Bengio, and G.~Hinton, ``Deep learning,'' \emph{nature}, vol. 521,
  no. 7553, p. 436, 2015.

\bibitem{krizhevsky2012imagenet}
A.~Krizhevsky, I.~Sutskever, and G.~E. Hinton, ``Imagenet classification with
  deep convolutional neural networks,'' in \emph{Advances in neural information
  processing systems}, 2012, pp. 1097--1105.

\bibitem{nam2016learning}
H.~Nam and B.~Han, ``Learning multi-domain convolutional neural networks for
  visual tracking,'' in \emph{Proceedings of the IEEE Conference on Computer
  Vision and Pattern Recognition}, 2016, pp. 4293--4302.

\bibitem{ciregan2012multi}
D.~Ciregan, U.~Meier, and J.~Schmidhuber, ``Multi-column deep neural networks
  for image classification,'' in \emph{Computer Vision and Pattern Recognition
  (CVPR), 2012 IEEE Conference on}.\hskip 1em plus 0.5em minus 0.4em\relax
  IEEE, 2012, pp. 3642--3649.

\bibitem{gidaris2015object}
S.~Gidaris and N.~Komodakis, ``Object detection via a multi-region and semantic
  segmentation-aware cnn model,'' in \emph{Proceedings of the IEEE
  International Conference on Computer Vision}, 2015, pp. 1134--1142.

\bibitem{hariharan}
B.~Hariharan, P.~Arbel{\'a}ez, R.~Girshick, and J.~Malik, ``Hypercolumns for
  object segmentation and fine-grained localization,'' in \emph{Proceedings of
  the IEEE conference on computer vision and pattern recognition}, 2015, pp.
  447--456.

\bibitem{hariharan1}
J.~Malik, P.~Arbel{\'a}ez, J.~Carreira, K.~Fragkiadaki, R.~Girshick,
  G.~Gkioxari, S.~Gupta, B.~Hariharan, A.~Kar, and S.~Tulsiani, ``The three
  r’s of computer vision: Recognition, reconstruction and reorganization,''
  \emph{Pattern Recognition Letters}, vol.~72, pp. 4--14, 2016.

\bibitem{deepsat}
S.~Basu, S.~Ganguly, S.~Mukhopadhyay, R.~DiBiano, M.~Karki, and R.~Nemani,
  ``Deepsat: a learning framework for satellite imagery,'' in \emph{Proceedings
  of the 23rd SIGSPATIAL International Conference on Advances in Geographic
  Information Systems}.\hskip 1em plus 0.5em minus 0.4em\relax ACM, 2015,
  p.~37.

\bibitem{ieeetran}
S.~Basu, S.~Ganguly, R.~R. Nemani, S.~Mukhopadhyay, G.~Zhang, C.~Milesi, A.~R.
  Michaelis, P.~Votava, R.~Dubayah, L.~Duncanson, B.~D. Cook, Y.~Yu,
  S.~Saatchi, R.~DiBiano, M.~Karki, E.~Boyda, U.~Kumar, and S.~Li, ``A
  semiautomated probabilistic framework for tree-cover delineation from 1-m
  {NAIP} imagery using a high-performance computing architecture,''
  \emph{{IEEE} Trans. Geoscience and Remote Sensing}, vol.~53, no.~10, pp.
  5690--5708, 2015.

\bibitem{urtasun1}
J.~Yao, S.~Fidler, and R.~Urtasun, ``Describing the scene as a whole: Joint
  object detection, scene classification and semantic segmentation,'' in
  \emph{Computer Vision and Pattern Recognition (CVPR), 2012 IEEE Conference
  on}.\hskip 1em plus 0.5em minus 0.4em\relax IEEE, 2012, pp. 702--709.

\bibitem{texture}
S.~Basu, M.~Karki, S.~Mukhopadhyay, S.~Ganguly, R.~R. Nemani, R.~DiBiano, and
  S.~Gayaka, ``A theoretical analysis of deep neural networks for texture
  classification,'' in \emph{2016 International Joint Conference on Neural
  Networks, {IJCNN} 2016, Vancouver, BC, Canada, July 24-29, 2016}, 2016, pp.
  992--999.

\bibitem{character}
S.~Basu, M.~Karki, S.~Ganguly, R.~DiBiano, S.~Mukhopadhyay, S.~Gayaka,
  R.~Kannan, and R.~R. Nemani, ``Learning sparse feature representations using
  probabilistic quadtrees and deep belief nets,'' \emph{Neural Processing
  Letters}, vol.~45, no.~3, pp. 855--867, 2017.

\bibitem{texture1}
S.~Basu, S.~Mukhopadhyay, M.~Karki, R.~DiBiano, S.~Ganguly, R.~R. Nemani, and
  S.~Gayaka, ``Deep neural networks for texture classification - {A}
  theoretical analysis,'' \emph{Neural Networks}, vol.~97, pp. 173--182, 2018.

\bibitem{he2016deep}
K.~He, X.~Zhang, S.~Ren, and J.~Sun, ``Deep residual learning for image
  recognition,'' in \emph{Proceedings of the IEEE conference on computer vision
  and pattern recognition}, 2016, pp. 770--778.

\bibitem{girshick2014rich}
R.~Girshick, J.~Donahue, T.~Darrell, and J.~Malik, ``Rich feature hierarchies
  for accurate object detection and semantic segmentation,'' in
  \emph{Proceedings of the IEEE conference on computer vision and pattern
  recognition}, 2014, pp. 580--587.

\bibitem{chen2018deeplab}
L.-C. Chen, G.~Papandreou, I.~Kokkinos, K.~Murphy, and A.~L. Yuille, ``Deeplab:
  Semantic image segmentation with deep convolutional nets, atrous convolution,
  and fully connected crfs,'' \emph{IEEE transactions on pattern analysis and
  machine intelligence}, vol.~40, no.~4, pp. 834--848, 2018.

\bibitem{tracking}
L.~Wang, W.~Ouyang, X.~Wang, and H.~Lu, ``Visual tracking with fully
  convolutional networks,'' in \emph{Proceedings of the IEEE International
  Conference on Computer Vision}, 2015, pp. 3119--3127.

\bibitem{Hastie}
T.~Hastie, R.~Tibshirani, and M.~Wainwright, \emph{Statistical Learning with
  Sparsity: The Lasso and Generalizations}.\hskip 1em plus 0.5em minus
  0.4em\relax Chapman \& Hall/CRC, 2015.

\bibitem{zeroshotRomero}
B.~Romera-Paredes and P.~Torr, ``An embarrassingly simple approach to zero-shot
  learning,'' in \emph{International Conference on Machine Learning}, 2015, pp.
  2152--2161.

\bibitem{ZeroshotNg}
R.~Socher, M.~Ganjoo, C.~D. Manning, and A.~Ng, ``Zero-shot learning through
  cross-modal transfer,'' in \emph{Advances in neural information processing
  systems}, 2013, pp. 935--943.

\bibitem{Oneshot}
L.~Fei-Fei, R.~Fergus, and P.~Perona, ``One-shot learning of object
  categories,'' \emph{IEEE transactions on pattern analysis and machine
  intelligence}, vol.~28, no.~4, pp. 594--611, 2006.

\bibitem{Vicarious}
``Advances in neural information processing systems 29: Annual conference on
  neural information processing systems 2016, december 5-10, 2016, barcelona,
  spain,'' D.~D. Lee, M.~Sugiyama, U.~von Luxburg, I.~Guyon, and R.~Garnett,
  Eds., 2016.

\bibitem{sabour2017dynamic}
S.~Sabour, N.~Frosst, and G.~E. Hinton, ``Dynamic routing between capsules,''
  in \emph{Advances in Neural Information Processing Systems}, 2017, pp.
  3859--3869.

\bibitem{Unsupervised}
M.~E. Celebi and K.~Aydin, \emph{Unsupervised Learning Algorithms},
  1st~ed.\hskip 1em plus 0.5em minus 0.4em\relax Springer Publishing Company,
  Incorporated, 2016.

\bibitem{sabahi2017hopfield}
F.~Sabahi, M.~O. Ahmad, and M.~Swamy, ``Hopfield network-based image retrieval
  using re-ranking and voting,'' in \emph{Electrical and Computer Engineering
  (CCECE), 2017 IEEE 30th Canadian Conference on}.\hskip 1em plus 0.5em minus
  0.4em\relax IEEE, 2017, pp. 1--4.

\bibitem{ren2015faster}
S.~Ren, K.~He, R.~Girshick, and J.~Sun, ``Faster r-cnn: Towards real-time
  object detection with region proposal networks,'' in \emph{Advances in neural
  information processing systems}, 2015, pp. 91--99.

\bibitem{pan2010}
S.~J. Pan and Q.~Yang, ``A survey on transfer learning,'' \emph{IEEE
  Transactions on knowledge and data engineering}, vol.~22, no.~10, pp.
  1345--1359, 2010.

\bibitem{yosinski2014transferable}
J.~Yosinski, J.~Clune, Y.~Bengio, and H.~Lipson, ``How transferable are
  features in deep neural networks?'' in \emph{Advances in neural information
  processing systems}, 2014, pp. 3320--3328.

\bibitem{deng2009imagenet}
J.~Deng, W.~Dong, R.~Socher, L.-J. Li, K.~Li, and L.~Fei-Fei, ``Imagenet: A
  large-scale hierarchical image database,'' in \emph{Computer Vision and
  Pattern Recognition, 2009. CVPR 2009. IEEE Conference on}.\hskip 1em plus
  0.5em minus 0.4em\relax IEEE, 2009, pp. 248--255.

\bibitem{karpathy2014large}
A.~Karpathy, G.~Toderici, S.~Shetty, T.~Leung, R.~Sukthankar, and L.~Fei-Fei,
  ``Large-scale video classification with convolutional neural networks,'' in
  \emph{Proceedings of the IEEE conference on Computer Vision and Pattern
  Recognition}, 2014, pp. 1725--1732.

\bibitem{tzeng2015simultaneous}
E.~Tzeng, J.~Hoffman, T.~Darrell, and K.~Saenko, ``Simultaneous deep transfer
  across domains and tasks,'' in \emph{Proceedings of the IEEE International
  Conference on Computer Vision}, 2015, pp. 4068--4076.

\bibitem{krotov2016dense}
D.~Krotov and J.~J. Hopfield, ``Dense associative memory for pattern
  recognition,'' in \emph{Advances in Neural Information Processing Systems},
  2016, pp. 1172--1180.

\bibitem{krotov2017dense}
------, ``Dense associative memory is robust to adversarial inputs,''
  \emph{arXiv preprint arXiv:1701.00939}, 2017.

\bibitem{he2015delving}
K.~He, X.~Zhang, S.~Ren, and J.~Sun, ``Delving deep into rectifiers: Surpassing
  human-level performance on imagenet classification,'' in \emph{Proceedings of
  the IEEE international conference on computer vision}, 2015, pp. 1026--1034.

\bibitem{mnih2015human}
V.~Mnih, K.~Kavukcuoglu, D.~Silver, A.~A. Rusu, J.~Veness, M.~G. Bellemare,
  A.~Graves, M.~Riedmiller, A.~K. Fidjeland, G.~Ostrovski \emph{et~al.},
  ``Human-level control through deep reinforcement learning,'' \emph{Nature},
  vol. 518, no. 7540, pp. 529--533, 2015.

\bibitem{girshick2015fast}
R.~Girshick, ``Fast r-cnn,'' in \emph{Proceedings of the IEEE international
  conference on computer vision}, 2015, pp. 1440--1448.

\bibitem{masci2011stacked}
J.~Masci, U.~Meier, D.~Cire{\c{s}}an, and J.~Schmidhuber, ``Stacked
  convolutional auto-encoders for hierarchical feature extraction,''
  \emph{Artificial Neural Networks and Machine Learning--ICANN 2011}, pp.
  52--59, 2011.

\bibitem{kingma2014semi}
D.~P. Kingma, S.~Mohamed, D.~J. Rezende, and M.~Welling, ``Semi-supervised
  learning with deep generative models,'' in \emph{Advances in Neural
  Information Processing Systems}, 2014, pp. 3581--3589.

\bibitem{gong2014compressing}
Y.~Gong, L.~Liu, M.~Yang, and L.~Bourdev, ``Compressing deep convolutional
  networks using vector quantization,'' \emph{arXiv preprint arXiv:1412.6115},
  2014.

\bibitem{ke2016fast}
S.~Ke, Y.~Zhao, B.~Li, Z.~Wu, and X.~Liu, ``Fast image clustering based on
  convolutional neural network and binary k-means,'' in \emph{Eighth
  International Conference on Digital Image Processing (ICDIP 2016)}, vol.
  10033.\hskip 1em plus 0.5em minus 0.4em\relax International Society for
  Optics and Photonics, 2016, p. 100332E.

\bibitem{wang2015semantic}
P.~Wang, J.~Xu, B.~Xu, C.~Liu, H.~Zhang, F.~Wang, and H.~Hao, ``Semantic
  clustering and convolutional neural network for short text categorization,''
  in \emph{Proceedings of the 53rd Annual Meeting of the Association for
  Computational Linguistics and the 7th International Joint Conference on
  Natural Language Processing (Volume 2: Short Papers)}, vol.~2, 2015, pp.
  352--357.

\bibitem{aleksander1984wisard}
I.~Aleksander, W.~Thomas, and P.~Bowden, ``Wisard⊙ a radical step forward in
  image recognition,'' \emph{Sensor review}, vol.~4, no.~3, pp. 120--124, 1984.

\bibitem{hopfield1984neurons}
J.~J. Hopfield, ``Neurons with graded response have collective computational
  properties like those of two-state neurons,'' \emph{Proceedings of the
  national academy of sciences}, vol.~81, no.~10, pp. 3088--3092, 1984.

\bibitem{george2017generative}
D.~George, W.~Lehrach, K.~Kansky, M.~L{\'a}zaro-Gredilla, C.~Laan, B.~Marthi,
  X.~Lou, Z.~Meng, Y.~Liu, H.~Wang \emph{et~al.}, ``A generative vision model
  that trains with high data efficiency and breaks text-based captchas,''
  \emph{Science}, vol. 358, no. 6368, p. eaag2612, 2017.

\bibitem{ladwani2017hopfield}
V.~M. Ladwani, Y.~Vaishnavi, R.~Shreyas, B.~V. Kumar, N.~Harisha, S.~Yogesh,
  P.~Shivaganga, and V.~Ramasubramanian, ``Hopfield net framework for audio
  search,'' in \emph{Communications (NCC), 2017 Twenty-third National
  Conference on}.\hskip 1em plus 0.5em minus 0.4em\relax IEEE, 2017, pp. 1--6.

\bibitem{fei2007learning}
L.~Fei-Fei, R.~Fergus, and P.~Perona, ``Learning generative visual models from
  few training examples: An incremental bayesian approach tested on 101 object
  categories,'' \emph{Computer vision and Image understanding}, vol. 106,
  no.~1, pp. 59--70, 2007.

\bibitem{griffin2007caltech}
G.~Griffin, A.~Holub, and P.~Perona, ``Caltech-256 object category dataset,''
  2007.

\bibitem{krizhevsky2009learning}
A.~Krizhevsky and G.~Hinton, ``Learning multiple layers of features from tiny
  images,'' 2009.

\bibitem{simonyan2014very}
K.~Simonyan and A.~Zisserman, ``Very deep convolutional networks for
  large-scale image recognition,'' \emph{arXiv preprint arXiv:1409.1556}, 2014.

\bibitem{bo2013multipath}
L.~Bo, X.~Ren, and D.~Fox, ``Multipath sparse coding using hierarchical
  matching pursuit,'' in \emph{Proceedings of the IEEE Conference on Computer
  Vision and Pattern Recognition}, 2013, pp. 660--667.

\bibitem{he2014spatial}
K.~He, X.~Zhang, S.~Ren, and J.~Sun, ``Spatial pyramid pooling in deep
  convolutional networks for visual recognition,'' in \emph{european conference
  on computer vision}.\hskip 1em plus 0.5em minus 0.4em\relax Springer, 2014,
  pp. 346--361.

\bibitem{snoek2015scalable}
J.~Snoek, O.~Rippel, K.~Swersky, R.~Kiros, N.~Satish, N.~Sundaram, M.~Patwary,
  M.~Prabhat, and R.~Adams, ``Scalable bayesian optimization using deep neural
  networks,'' in \emph{International Conference on Machine Learning}, 2015, pp.
  2171--2180.

\bibitem{zeiler2014visualizing}
M.~D. Zeiler and R.~Fergus, ``Visualizing and understanding convolutional
  networks,'' in \emph{European conference on computer vision}.\hskip 1em plus
  0.5em minus 0.4em\relax Springer, 2014, pp. 818--833.

\bibitem{kuang2015hardware}
D.~Kuang, A.~Gittens, and R.~Hamid, ``Hardware compliant approximate image
  codes,'' in \emph{Proceedings of the IEEE Conference on Computer Vision and
  Pattern Recognition}, 2015, pp. 924--932.

\bibitem{shaban2013local}
A.~Shaban, H.~R. Rabiee, M.~Farajtabar, and M.~Ghazvininejad, ``From local
  similarity to global coding: An application to image classification,'' in
  \emph{Proceedings of the IEEE Conference on Computer Vision and Pattern
  Recognition}, 2013, pp. 2794--2801.

\bibitem{singh2017scatternet}
A.~Singh and N.~Kingsbury, ``Scatternet hybrid deep learning (shdl) network for
  object classification,'' \emph{arXiv preprint arXiv:1708.09212}, 2017.

\bibitem{zhu2014submodular}
F.~Zhu, Z.~Jiang, and L.~Shao, ``Submodular object recognition,'' in
  \emph{Proceedings of the IEEE Conference on Computer Vision and Pattern
  Recognition}, 2014, pp. 2457--2464.

\bibitem{springenberg2014striving}
J.~T. Springenberg, A.~Dosovitskiy, T.~Brox, and M.~Riedmiller, ``Striving for
  simplicity: The all convolutional net,'' \emph{arXiv preprint
  arXiv:1412.6806}, 2014.

\bibitem{lee2016generalizing}
C.-Y. Lee, P.~W. Gallagher, and Z.~Tu, ``Generalizing pooling functions in
  convolutional neural networks: Mixed, gated, and tree,'' in \emph{Artificial
  Intelligence and Statistics}, 2016, pp. 464--472.

\bibitem{radford2015unsupervised}
A.~Radford, L.~Metz, and S.~Chintala, ``Unsupervised representation learning
  with deep convolutional generative adversarial networks,'' \emph{arXiv
  preprint arXiv:1511.06434}, 2015.

\bibitem{exemplar1}
A.~Dosovitskiy, P.~Fischer, J.~T. Springenberg, M.~Riedmiller, and T.~Brox,
  ``Discriminative unsupervised feature learning with exemplar convolutional
  neural networks,'' \emph{IEEE Transactions on Pattern Analysis and Machine
  Intelligence}, vol.~38, no.~9, pp. 1734--1747, Sept 2016.

\bibitem{catg}
J.~T. Springenberg, ``Unsupervised and semi-supervised learning with
  categorical generative adversarial networks,'' \emph{arXiv preprint
  arXiv:1511.06390}, 2015.

\bibitem{mcdonnell2015enhanced}
M.~D. McDonnell and T.~Vladusich, ``Enhanced image classification with a
  fast-learning shallow convolutional neural network,'' in \emph{Neural
  Networks (IJCNN), 2015 International Joint Conference on}.\hskip 1em plus
  0.5em minus 0.4em\relax IEEE, 2015, pp. 1--7.

\bibitem{chan2015pcanet}
T.-H. Chan, K.~Jia, S.~Gao, J.~Lu, Z.~Zeng, and Y.~Ma, ``Pcanet: A simple deep
  learning baseline for image classification?'' \emph{IEEE Transactions on
  Image Processing}, vol.~24, no.~12, pp. 5017--5032, 2015.

\bibitem{Krizhevsky:2014:Misc}
A.~Krizhevsky, ``cuda-convnet,''
  \url{https://code.google.com/archive/p/cuda-convnet/}, July 18 2014.

\bibitem{mairal2014convolutional}
J.~Mairal, P.~Koniusz, Z.~Harchaoui, and C.~Schmid, ``Convolutional kernel
  networks,'' in \emph{Advances in neural information processing systems},
  2014, pp. 2627--2635.

\end{thebibliography}

\end{document}